\documentclass[letterpaper, 10 pt, conference]{ieeeconf}  % Comment this line out if you need a4paper

\IEEEoverridecommandlockouts                              % This command is only needed if 
\usepackage{graphics} % for pdf, bitmapped graphics files
\usepackage{booktabs}
\usepackage{multirow}
\usepackage{graphicx}
\usepackage{epsfig} % for postscript graphics files
\usepackage{mathptmx} % assumes new font selection scheme installed
\usepackage{times} % assumes new font selection scheme installed
\usepackage{amsmath} % assumes amsmath package installed
\usepackage{amssymb}  % assumes amsmath package installed
\usepackage{eucal}
\usepackage{booktabs}
\usepackage{adjustbox}
\usepackage{subcaption}
\usepackage{algorithm}
\usepackage{algpseudocode}
\usepackage{hyperref}
\usepackage{gensymb}
\usepackage[table]{xcolor}
\usepackage{cite}

\def\BibTeX{{\rm B\kern-.05em{\sc i\kern-.025em b}\kern-.08em
    T\kern-.1667em\lower.7ex\hbox{E}\kern-.125emX}}

\usepackage{amsmath}
\usepackage{amssymb}
\usepackage{mathtools} % for prescript
\usepackage{bm}

\newcommand{\lie}[1]{\mathcal{L}_{#1}}

\newcommand{\state}[0]{\bm{x}}

\newcommand{\action}[0]{\bm{u}}
\newcommand{\naction}[0]{m}

\newcommand{\ostate}[0]{\bm{x}_o}

\newcommand{\dostate}[0]{\dot{\bm{x}}_o}

\title{\LARGE \bf
Memory-Aware Multi-Sensor Perception for Efficient and Safe Navigation in Dynamic Environments
}

\author{Jingshuo Li, Yifan Xue, Yifei Li, Shubhodeep Shiv Aditya, and Nadia Figueroa
\thanks{The authors are with the School of Engineering and Applied Science, University of Pennsylvania, Philadelphia, USA (e-mail:jingshuo@alumni.upenn.edu, \{yifanxue, liyf, nadiafig\}@engineering.upenn.edu). }}

\begin{document}
\definecolor{Gray}{gray}{0.85}
\definecolor{TableOverall}{gray}{0.92}
\maketitle
\thispagestyle{empty}
\pagestyle{empty}

%%%%%%%%%%%%%%%%%%%%%%%%%%%%%%%%%%%%%%%%%%%%%%%%%%%%%%%%%%%%%%%%%%%%%%%%%%%%%%%%
\begin{abstract}

Autonomous navigation in previously unseen environments requires effective perception, persistent environmental representation, and collision avoidance while maintaining progress toward a goal. Existing perception-based methods often rely on prior maps or short-horizon observations, limiting their ability to exploit previously observed structure. We propose a memory-aware multi-sensor navigation framework that integrates LiDAR and RGB perception, online distance-field representation learning, and a stage-adaptive Modulated Control Barrier Function Quadratic Program (MCBF-QP). The framework persistently represents static infrastructure while tracking dynamic obstacles, enabling the MCBF-QP controller to exploit previously observed geometry for obstacle circumvention and adapt its safety constraints and guidance to local conditions. Experiments in complex indoor and outdoor environments demonstrate improved navigation efficiency and goal-reaching performance while maintaining collision avoidance in narrow passages and around dynamic obstacles. The project website, including videos and supplementary materials, is available at \url{https://yifanxueseas.github.io/memory-aware-multi-sensory-navigation-web/}. 

\end{abstract}
\begin{figure*}[t]
    \centering
    \includegraphics[
        width=0.9\linewidth,
        trim=0mm 6mm 0mm 3cm,
        clip
    ]{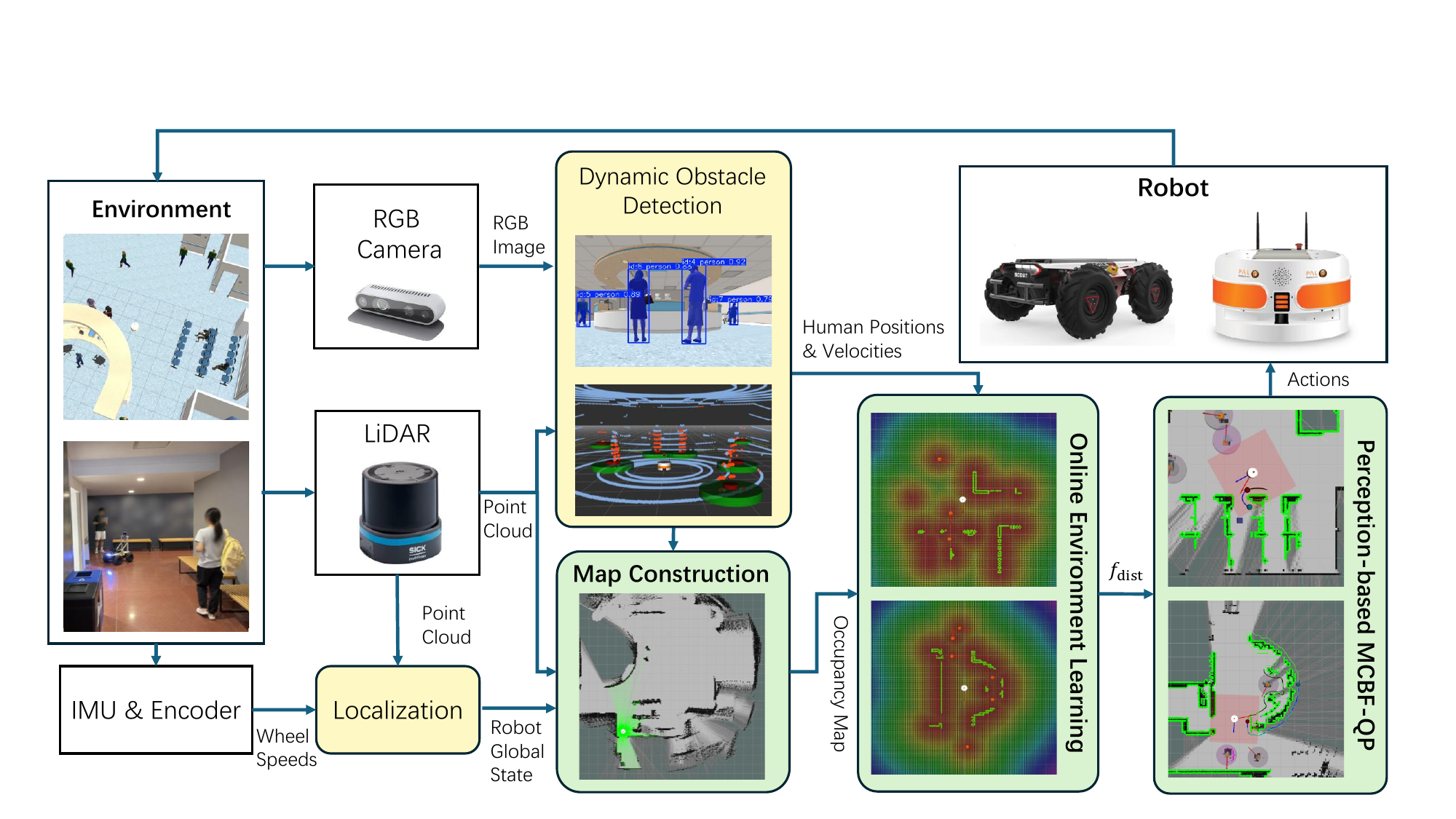}
    \caption{Overview of the proposed \textbf{memory-aware multi-sensory perception and navigation framework}, integrating onboard state estimation, dynamic obstacle detection, online environment representation learning, and stage-adaptive MCBF control for safe and live navigation in previously unmapped environments.}
    \vspace{-13pt}

    \label{fig:block diagram}
\end{figure*}

\section{Introduction}
Recent advances in autonomous systems have driven substantial progress in robot navigation and collision avoidance. A broad spectrum of approaches has demonstrated strong empirical performance, including graph-search methods, data-driven approaches, sampling-based planners, closed-form reactive strategies, and control optimization \cite{lavalle1998rapidly, hart1968formal, dolgov2008practical,lutjens2019safe, zhu2022collision, hope,williams2017model, khansari2012dynamical, LukesDS, onManifoldMod, mpcdc2021safety, admm_mcbf_lca, xue2026proactivelocalminimafreerobotnavigation}. However, deploying these approaches in previously unseen environments remains challenging because many navigation and safe-control methods rely on prior knowledge of the environment, such as prebuilt maps or externally provided observations. To relax this requirement, perception-based navigation pipelines have increasingly leveraged onboard sensors, including LiDARs and RGB-D cameras, to construct or infer environmental representations online \cite{slam_Dijkstra_av,deep_learning_lidar_gps_av, mpc_cbf_ellipse_lidar, teb_lidar_dynamic, cbf_circulation, cbf_nerf_camera,rnbf}.

Existing perception-based approaches can be broadly distinguished by whether they retain an explicit representation of previously observed environments. Some methods perform long-term navigation using SLAM together with pre-scanned or GPS-referenced maps \cite{slam_Dijkstra_av, deep_learning_lidar_gps_av}. While such approaches can exploit global environmental structure, they depend on the availability of a suitable prior map and can become unreliable when the environment changes substantially. Other approaches construct only temporary local representations from onboard sensors to enable reactive path planning in dynamic environments, using either LiDAR \cite{mpc_cbf_ellipse_lidar, teb_lidar_dynamic, cbf_circulation} or RGB-D cameras \cite{cbf_nerf_camera,camera_rgbd_dynamic,rnbf}. These approaches avoid dependence on prior maps and enable fast local reactions, but typically discard useful geometric information once it leaves the local sensing horizon. As a result, the robot may repeatedly react to locally perceived obstacles without exploiting the structure of previously explored free space, increasing the likelihood of undesirable equilibria, or repeated exploration of already visited regions. We refer to the ability to maintain progress toward the goal while avoiding such behaviors as navigation \textbf{liveness}. Persistent representations of previously observed static infrastructure can therefore provide valuable information for improving navigation in the current environment and for subsequent tasks performed in the same environment.

Constructing such persistent representations using only onboard sensing requires robust geometric and semantic perception. As navigation tasks increasingly involve complex and dynamic environments, recent research has increasingly explored LiDAR-camera fusion to combine the complementary strengths of individual sensing modalities~\cite{lidar_camera_detection,lidar_camera_learning}. LiDARs provide accurate, long-range geometric measurements with wide angular coverage, while RGB cameras provide rich visual and semantic information for identifying objects and environmental structures~\cite{camera_rgbd_dynamic}. Nevertheless, existing LiDAR-camera fusion pipelines have primarily demonstrated their capabilities in structured environments and have yet to address navigation in complex, unstructured indoor environments that require traversing multiple rooms, reasoning about persistent spatial structure, and avoiding dynamic agents such as pedestrians.

Persistent environmental representation alone, however, does not guarantee successful navigation. The navigation method must also exploit the available environmental information while maintaining both collision avoidance and progress toward the goal. Existing perception-based navigation pipelines have employed a variety of approaches, including deep learning, Timed Elastic Band (TEB), Model Predictive Control (MPC), and Control Barrier Function (CBF)-based controllers \cite{deep_learning_lidar_gps_av, teb_lidar_dynamic, mpc_cbf_ellipse_lidar, cbf_circulation}. While these methods have demonstrated effective navigation in various environments, they generally do not provide explicit guarantees of convergence without undesirable equilibria or dysfunctional circulation. Recent work on Modulated Control Barrier Function Quadratic Programs (MCBF-QPs) addresses this limitation by combining collision avoidance with navigation \emph{liveness}, i.e., continued progress toward the goal without undesirable equilibria \cite{xue_2025_mcbf,admm_mcbf_lca,xue2026proactivelocalminimafreerobotnavigation}.

% Building on the MCBF-QP framework, we propose a multi-sensor navigation system that is \emph{memory-aware}: rather than relying solely on instantaneous observations, the robot retains geometric information about previously observed static infrastructure and incorporates it into subsequent navigation decisions. LiDAR and RGB-D observations are jointly used to learn a distance-field-based representation of the explored environment online while distinguishing dynamic obstacles from persistent infrastructure. The accumulated representation is then exploited by the MCBF-QP controller to construct obstacle-circumventing guidance, enabling more efficient navigation through complex environments without requiring an offline map.

These observations motivate a \textbf{memory-aware} navigation framework that leverages previously observed environmental structure, rather than relying solely on instantaneous perception, while providing \textbf{liveness-aware} collision avoidance, without requiring an offline map.
Thus, the main contributions of this paper are as follows:
\begin{enumerate}
\item We propose a \textbf{multi-sensor} perception and navigation framework for autonomous navigation in complex, unknown, and unstructured indoor environments.
\item We develop an online distance-field-based environmental representation that \textbf{persistently} captures static infrastructure while distinguishing dynamic obstacles, without requiring an offline map.
\item We integrate the persistent representation with an MCBF-QP-based controller to exploit previously observed environmental structure for obstacle circumvention, and demonstrate \textbf{improved mapless navigation} efficiency and goal-reaching performance in dynamic environments.
\end{enumerate}

\section{Preliminaries}
\subsection{Assumptions}
\label{sec:model&assumptions}
We consider a robotic system governed by the control-affine dynamics
\begin{equation}
    \dot{\bm{\state}}=f(\state)+g(\state)\action,
    \label{eq:system_dynamics}
\end{equation}
where $\state \in \mathbb{R}^n$ is the robot state, $\action \in \mathbb{R}^m$ is the control action, and $f$ and $g$ are Lipschitz continuous functions. 

% For the navigation environment, we assume that large static structures that partition the navigable space can be detected and approximated as vertical planar surfaces, with surface normals approximately orthogonal to the gravity direction. Common examples include walls in indoor environments and building facades or other large structures in outdoor environments.

\subsection{Gaussian Process Distance Field}
\label{sec:gpdf}

In the proposed pipeline, we use a Gaussian Process Distance Field (GPDF)~\cite{gpdf,le_2023_gpdf} to obtain a continuous and differentiable distance representation from observed obstacle boundaries. Given a set of obstacle boundary points
$\mathcal{P}=\{\bm{p}_i\}$, let $o(\bm{p})$ denote a latent field modeled as a Gaussian process (GP),
\begin{equation}
    o(\bm{p}) \sim
    \mathcal{GP}\!\left(0,k_o(\bm{p},\bm{p}')\right),
\end{equation}
where $k_o$ is selected such that its inverse transformation $f_{\mathrm{inv}}$ recovers the Euclidean distance:
\begin{equation}
    f_{\mathrm{inv}}
    \left(k_o(\bm{p},\bm{p}')\right)
    \coloneqq
    \|\bm{p}-\bm{p}'\|_2.
    \label{eq:gpdf_inverse}
\end{equation}
Assigning a unit latent observation to each boundary point, GP regression gives the posterior mean
\begin{equation}
    \bar{o}(\bm{p})
    =
    k_o(\bm{p},\mathcal{P})
    \left(K_o^\mathcal{P}+\sigma_o^2\mathbf{I}\right)^{-1}
    \mathbf{1},
    \label{eq:gpdf_mean}
\end{equation}
where $K_o^\mathcal{P}\coloneqq K_o(\mathcal{P},\mathcal{P})$ is the covariance matrix evaluated at the boundary points, $\sigma_o$ is the observation-noise parameter, and $\mathbf{1}$ is a vector of unit observations. The resulting distance estimate is obtained by applying the inverse transformation to the posterior mean:
\begin{equation}
    f_{\mathrm{dist}}(\bm{p})
    =
    f_{\mathrm{inv}}\!\left(\bar{o}(\bm{p})\right).
    \label{eq:gpdf_predicted_distance}
\end{equation}
Since both $k_o$ and $f_{\mathrm{inv}}$ are differentiable, $f_{\mathrm{dist}}$ is differentiable with an analytically computable gradient, making it suitable for constructing the differentiable barrier functions used by the controller.

\subsection{Modulated Control Barrier Functions}
\label{sec:mcbf}
To retain CBF-QP safety while enhancing \textbf{liveness}, we adopt the on-manifold Modulated CBF-QP (MCBF-QP) as the controller, which provides proven safety and convergence guarantees. MCBF-QP augments the CBF constraint with a tangent-guidance constraint that enforces positive velocity along an estimated obstacle boundary geodesic guiding vector $\phi(\state)$, thereby steering the robot along an obstacle-circumventing direction and mitigating undesirable equilibria and dysfunctional circulation~\cite{xue_2025_mcbf,xue2026proactivelocalminimafreerobotnavigation}.

For multiple nearby obstacles, we modify the original MCBF-QP by replacing its unified barrier function with individual barrier functions $h_o(\state,\ostate)$ for each obstacle $o\in\mathcal{O}$. This modification imposes independent safety constraints for each obstacle cluster, yielding a more conservative representation that better preserves clearance from nearby obstacles. The resulting formulation is
\begin{align}
\nonumber
    \action_{\mathrm{mcbf}}
    &=\arg\min_{\action\in\mathbb{R}^{\naction}}
    (\action-\action_{\mathrm{nom}})^\top
    (\action-\action_{\mathrm{nom}}),
    \nonumber\\
    \text{s.t.}\qquad
    &\lie{f}h_o
    +\lie{g}h_o\,\action
    +\nabla_{\ostate}h_o^\top\dostate
    \geq
    -\alpha(h_o)
    \quad \forall o\in\mathcal{O},
    \label{eq:cbf-qp safety constraints}
\end{align}
\begin{align}
    % \text{s.t.}\qquad
    % &\lie{f}h_o
    % +\lie{g}h_o\,\action
    % +\nabla_{\ostate}h_o^\top\dostate
    % \geq
    % -\alpha(h_o)
    % \quad \forall o\in\mathcal{O},
    % \label{eq:cbf-qp safety constraints}\\
    &\phi(\state)^\top f(\state)
    +\phi(\state)^\top g(\state)\action
    \geq
    \nu_\phi.
    \label{eq:mod_phi_cbf_constraint_affine}
\end{align}
Here, $\ostate$ and $\dostate$ denote the state and velocity of obstacle $o$, respectively, $\alpha(\cdot)$ is a class-$\mathcal{K}$ function, and $\nu_\phi>0$ is a user-defined constant specifying the minimum velocity component along $\phi$. $\lie{f},\lie{g}$ denote the Lie derivatives wrt. each component of the control-affine dynamics in Eq.~\ref{eq:system_dynamics}. 

\section{Methodology}
As shown in Fig.~\ref{fig:block diagram}, the proposed navigation framework integrates onboard state estimation, multi-sensory perception, online environment representation learning, and stage-adaptive MCBF control to enable collision avoidance and enhance navigation liveness in previously unmapped environments. Onboard state estimation establishes a persistent reference frame, while multi-sensory perception identifies dynamic obstacles and environment geometry from LiDAR and RGB observations (Sections~\ref{sec:state_estimation} and~\ref{sec:dynamic_obs}). The resulting environmental observations are incorporated into an online-learned distance-field representation (Section~\ref{sec:mapping}), which enables the MCBF controller to enforce collision-avoidance constraints and generate obstacle-circumventing guidance. The controller further adapts its CBF constraints and geodesic guidance to the local navigation condition, tailoring its behavior to dynamic obstacles, narrow passages, and open space (Section~\ref{sec:adaptive_controls}).

\subsection{Onboard State Estimation}
\label{sec:state_estimation}

To maintain a persistent environmental representation without an external motion-capture system, sensor observations and the robot state $\state$ are expressed in a common reference frame. The origin of the constructed frame is initialized at the robot's initial position, with the $z$-axis aligned with the gravity direction and the $x$-$y$ axes spanning the horizontal plane. The localization pipeline combines the 6-DOF LiDAR--inertial pose from FAST-LIO2~\cite{fastlio2} with longitudinal wheel velocity through an extended Kalman filter~\cite{robotlocalization}. LiDAR--inertial odometry provides spatial pose estimation, while wheel odometry provides complementary short-term motion information when LiDAR scan registration is degraded. The fused pose and velocity define the robot state $\state$ in the reference frame and are used to transform LiDAR and camera observations into the same frame for persistent mapping. All mathematical quantities in the following sections are expressed in this reference frame unless otherwise specified.

\subsection{Dynamic Obstacle Detection}
\label{sec:dynamic_obs}
Independently movable obstacles, whether in motion or not, pose a higher safety risk than static obstacles and require accurate position and velocity estimates to maintain sufficient clearance and avoid collisions. LiDAR-based dynamic obstacle detection is typically geometry-based~\cite{przybyla2017detection,teb_lidar_dynamic}, but geometric cues alone cannot reliably distinguish movable obstacles from similarly shaped static infrastructure or separate a person from nearby obstacles. In contrast, the YOLO instance-segmentation model~\cite{ultralytics2026yolo26} provides semantic identification but not reliable metric position or velocity estimates. LiDAR and RGB observations are therefore jointly adopted in the proposed pipeline, with the two detection branches running in parallel to support accurate identification and tracking of such obstacles.

The LiDAR branch filters $\mathcal{P}_{\mathrm{lidar}}$ along the $z$-axis to remove floor points and structures well above the robot. The remaining points are clustered and approximated as lines or circles, with line-like clusters treated as likely static infrastructure and circle-like clusters as candidate dynamic obstacles. Their centers are denoted by $\bm{p}_{c,k}^{\mathrm{geo}}\in\mathbb{R}^2$.

The camera branch processes RGB images using the YOLO instance-segmentation model~\cite{ultralytics2026yolo26}, producing an instance mask $M_{\mathrm{dyn}}^{(k)}(i,j)$ for each detected instance $k$, where $(i,j)$ denotes the pixel index. LiDAR points are projected into each image using the corresponding LiDAR-to-camera extrinsic transformation and camera intrinsic matrix. For each instance $k$, let $\mathcal{I}_k$ denote the set of pixel indices belonging to its instance mask,
\begin{equation}
\mathcal{I}_k
=
\left\{
(i,j):
M_{\mathrm{dyn}}^{(k)}(i,j)=1
\right\}.
\end{equation}
The corresponding mask centroid is
\begin{equation}
(i_c,j_c)
=
\frac{1}{|\mathcal{I}_k|}
\sum_{(i,j)\in\mathcal{I}_k}(i,j).
\end{equation}
Let $\mathcal{P}_M^{(k)}$ denote the LiDAR points whose projections lie within $M_{\mathrm{dyn}}^{(k)}$. The point whose projected pixel is closest to $(i_c,j_c)$ is derived as
\begin{equation}
\bm{p}_{c,k}^{\mathrm{image}}
=
\underset{\bm{p}\in\mathcal{P}_M^{(k)}}{\arg\min}
\left[
\left(i(\bm{p})-i_c\right)^2
+
\left(j(\bm{p})-j_c\right)^2
\right],
\end{equation}
where $(i(\bm{p}),j(\bm{p}))$ denotes the pixel coordinates of LiDAR point $\bm{p}$ in the RGB image. The mask centroid is used rather than the centroid of the enclosed LiDAR points because the segmentation mask may include points from nearby objects that are close in image space but have substantially different image depths.

The two detection results are then cross-validated for \emph{false-positive suppression} and \emph{false-negative recovery}. A geometric candidate $\bm{p}_{c,k}^{\mathrm{geo}}$ is retained only if a sufficient fraction of its neighboring LiDAR points project within a dynamic-obstacle mask $M_{\mathrm{dyn}}^{(k)}$, thereby rejecting geometrically similar static objects. Conversely, dynamic obstacles indicated by $M_{\mathrm{dyn}}^{(k)}$ but missed by geometric extraction are recovered when they cannot be distinguished from nearby static obstacles using LiDAR geometry alone. For each recovered detection, $\bm{p}_{c,k}^{\mathrm{image}}$ is added as its dynamic-obstacle position. The retained geometric detections and recovered image-based detections together form $\mathcal{P}_{\mathrm{dyn}}$.

Finally, $\mathcal{P}_{\mathrm{dyn}}$ is provided to the obstacle velocity estimator of~\cite{przybyla2017detection} to obtain $\dot{\bm{p}}_{c,k}$. The resulting multisensory detection pipeline outputs tracked dynamic obstacles represented by
\begin{equation}
\label{eq:obstacle_states}
s_k=
\left(
\bm{p}_{c,k},
\dot{\bm{p}}_{c,k}
\right).
\end{equation}

\subsection{Occupancy Grid Map Construction}
\label{sec:occupancy_map}

The occupancy grid mapping module constructs a 2D representation of the surrounding free and occupied space using ray tracing. Standard ray-tracing updates can leave stale occupied cells after a dynamic obstacle moves away, unnecessarily restricting the feasible control space and potentially resulting in infeasible or suboptimal actions. Tracked dynamic-obstacle states from Eq.~\eqref{eq:obstacle_states} in Sec.~\ref{sec:dynamic_obs} are therefore incorporated to explicitly remove vacated occupancy.

The incoming 3D LiDAR point cloud $\mathcal{P}_{\mathrm{lidar}}$ is first filtered along the $z$-axis to exclude floor and ceiling points and projected onto the horizontal plane as $\mathcal{P}_{\mathrm{lidar}}^{xy}$. The projected points are discretized into $N$ beams over $360^\circ$, with traversed cells identified using Bresenham's line algorithm~\cite{bresenham1965algorithm}. The occupancy grid is initialized with $\ell_{\mathrm{init}}=0$, and free and endpoint cells are updated with $\ell_{\mathrm{free}}<0$ and $\ell_{\mathrm{occ}}>0$, respectively, following the binary Bayesian occupancy formulation~\cite{elfes1989occupancy}. The log-odds values are bounded within $[\ell_{\min},\ell_{\max}]$, and cells with $\ell>\ell_{\mathrm{thr}}$ are classified as occupied.

For each tracked dynamic obstacle $k$, a circular mask centered at its estimated position $\bm{p}_{c,k}$ with radius $r_k$ is constructed at each update cycle. Cells within a dynamic-obstacle mask in the previous cycle but outside all such masks in the current cycle are identified as vacated and reset to $\ell_{\min}$. This targeted reset accelerates removal of vacated dynamic occupancy while preserving standard occupancy updates for static obstacles.

\subsection{Online Environment Representation Learning}
\label{sec:mapping}

This module integrates the occupancy grid and tracked dynamic-obstacle states to construct continuous obstacle representations for use by the MCBF-QP controller. The processing consists of three stages: (i) obstacle clustering, (ii) dynamic-obstacle separation, and (iii) distance-field learning. Obstacle clusters are first extracted from the occupancy grid maintained as described in Section~\ref{sec:occupancy_map}, then separated according to their association with tracked dynamic obstacles or environment infrastructure, and finally converted into suitable distance-field representations.

\subsubsection{Obstacle Clustering}
Occupied grid cells within a radius $r_{\mathrm{proc}}$ of the robot are first retained to limit the computational cost of subsequent processing while preserving sufficient environmental context for navigation toward the goal. These cells are clustered using density-based spatial clustering of applications with noise (DBSCAN)~\cite{ester1996dbscan}, with obstacle connectivity determined according to the robot's effective volume. For each cluster, a boundary contour is extracted from its binary occupancy mask using the marching squares algorithm~\cite{maple2003marching}. This boundary representation removes redundant interior cells while preserving the obstacle geometry, which is important for subsequent preprocessing before distance-field learning. The contours are converted from grid indices to $xy$ coordinates and collected as
\begin{equation}
\mathbb{P}
=
\left\{
\mathcal{P}_{j}^{(xy)}
\right\},
\end{equation}
where $\mathcal{P}_{j}^{(xy)}$ denotes the 2D boundary point set of the $j$-th cluster.

\subsubsection{Dynamic Obstacle Separation}

The tracked dynamic-obstacle states $\{s_k\}$ from Sec.~\ref{sec:dynamic_obs} are used to separate dynamic obstacles from environment infrastructure. For each boundary point set $\mathcal{P}_{j}^{(xy)}\in\mathbb{P}$, points within a radius $r_{\mathrm{dyn}}$ of any tracked dynamic obstacle are assigned to $\mathbb{P}_{\mathrm{dyn}}$, while the remaining points are assigned to $\mathbb{P}_{\mathrm{env}}$. Specifically,
\begin{align}
\nonumber
\mathbb{P}_{\mathrm{dyn}}
&=
\bigcup_j
\left\{
\bm{p}^{(xy)}\in\mathcal{P}_{j}^{(xy)}
\;\middle|\;
\min_k
\|\bm{p}^{(xy)}-\bm{p}_{c,k}\|_2
\leq r_{\mathrm{dyn}}
\right\},\\
\nonumber
\mathbb{P}_{\mathrm{env}}
&=
\bigcup_j
\left\{
\bm{p}^{(xy)}\in\mathcal{P}_{j}^{(xy)}
\;\middle|\;
\min_k
\|\bm{p}^{(xy)}-\bm{p}_{c,k}\|_2
> r_{\mathrm{dyn}}
\right\}.
\end{align}
Clusters containing fewer than $N_{\min}$ points after this separation are discarded as noise.

\subsubsection{Distance Field Learning}
The obstacle point sets, $\mathcal{P}_{j}^{(xy)}$, are then converted into distance fields for use by the MCBF-QP controller in Section.~\ref{sec:mcbf}. To enable real-time computation, GPDF construction and evaluation are implemented in JAX with just-in-time (JIT) compilation. Since the GP covariance matrix dimensions depend on the point-set cardinality, varying point-set sizes prevent reuse of the JIT-compiled routines and incur repeated compilation overhead. Each point set is therefore resampled to a fixed cardinality $n_s$:
\begin{equation}
\nonumber
\mathcal{P}_j^{(xy)} =
\begin{cases}
  \mathrm{SampleUniform}(\mathcal{P}_j^{(xy)},n_s), & |\mathcal{P}_j^{(xy)}| > n_s, \\
  \mathcal{P}_j^{(xy)} \cup
  \mathrm{Jitter}(\mathcal{P}_j^{(xy)},n_s-|\mathcal{P}_j^{(xy)}|),
  & |\mathcal{P}_j^{(xy)}| < n_s, \\
  \mathcal{P}_j^{(xy)}, & |\mathcal{P}_j^{(xy)}| = n_s,
\end{cases}
% \label{eq:resample}
\end{equation}
after removing coincident points. For oversized point sets, $\mathrm{SampleUniform}$ uniformly samples $n_s$ points from the boundary-only set $\mathcal{P}_j^{(xy)}$, avoiding interior points that can dominate when $|\mathcal{P}_j^{(xy)}|\gg n_s$ and distort the sampled obstacle geometry.  For undersized point sets, $\mathrm{Jitter}$ generates the required additional points by sampling $n_s-|\mathcal{P}_j^{(xy)}|$ existing boundary points and perturbing them with additive Gaussian noise~\cite{qi2017pointnet++}.

All obstacle point sets in $\mathbb{P}_{\mathrm{env}}$ and those in $\mathbb{P}_{\mathrm{dyn}}$ with sufficient spatial extent are represented by GPDFs, with a separate GPDF constructed for each obstacle \cite{gpdf}. For an obstacle point set $\mathcal{P}_j^{(xy)}$, the covariance between two points $\bm{p}^{(xy)}$ and $\bm{p}^{(xy)}_q$ is defined by the exponential kernel
\begin{equation}
k_o(\bm{p}^{(xy)},\bm{p}^{(xy)}_q)
=
\exp\left(
-\frac{\|\bm{p}^{(xy)}-\bm{p}^{(xy)}_q\|_2}{L}
\right),
\label{eq:gpdf_kernel}
\end{equation}
where $L$ determines the spatial decay of the covariance. At a query position $\bm{p}^{(xy)}$, the predicted latent field $\bar{o}(\bm{p}^{(xy)})$ is transformed into a distance field as
\begin{equation}
f_{\mathrm{dist}}(\bm{p}^{(xy)})
=
-L\log\left(\bar{o}(\bm{p}^{(xy)})\right)
+\mathrm{offset},
\label{eq:gpdf_dist}
\end{equation}
where a small $\mathrm{offset}$ is introduced to improve numerical stability near obstacle boundaries.

For small-scale dynamic obstacles, such as humans, sparse LiDAR provides insufficient boundary information for accurate GPDF estimation. These obstacles are therefore represented as circles using the estimated center position $\bm{p}_{c,k}$ and radius $r_k$ from Section.~\ref{sec:dynamic_obs}. The corresponding distance field at a query position $\bm{p}^{(xy)}$ is
\begin{equation}
f_{\mathrm{dist}}(\bm{p}^{(xy)})
=
\left\|
\bm{p}^{(xy)}-\bm{p}_{c,k}
\right\|_2-r_k.
\end{equation}

\subsection{Perception-Based Adaptive MCBF-QP}
\label{sec:adaptive_controls}
The proposed MCBF controller adapts its collision-avoidance constraints and geodesic guidance to the local navigation condition. The barrier function and geodesic guidance are first formulated from the perceived environment, after which the local condition determines the collision-avoidance stage and corresponding controller parameters.

\subsubsection{Perception-Based MCBF Formulation}

The barrier function for CBF-based collision avoidance is formulated as
\begin{equation}
    h(\state)
    =
    f_{\mathrm{dist}}(\bm{p})
    +
    f_{\mathrm{other}}(\state)
    +
    \delta,
\end{equation}
where $f_{\mathrm{other}}(\state)$ accounts for state-dependent terms not captured by the distance field, including non-Euclidean state variables, and $\delta$ is the safety margin. With the safe set defined by $\{\state:h(\state)\geq0\}$, the barrier-function derivative is constrained by
$\dot{h}(\state,\action)\geq-\alpha(h(\state))$,
as in Eq.~\eqref{eq:cbf-qp safety constraints}. The barrier-function derivative associated with environment infrastructure is
\begin{equation}
    \nonumber 
    \dot{h}_{\mathrm{env}}
    =
    \nabla_{\bm{p}}f_{\mathrm{dist}}^\top
    \left(
        f_p(\state)+g_p(\state)\action
    \right)
    +
    \nabla_{\state}f_{\mathrm{other}}^\top
    \left(
        f(\state)+g(\state)\action
    \right),
\end{equation}
while the motion of dynamic obstacle $k$, with estimated velocity
$\dot{\bm{p}}_{c,k}$, introduces the relative-velocity term
\begin{align}
\nonumber
    \dot{h}_{\mathrm{dyn}}
    &=
    \nabla_{\bm{p}}f_{\mathrm{dist}}^{(k)\top}
    \left(
        f_p(\state)+g_p(\state)\action-\dot{\bm{p}}_{c,k}
    \right)\\
    &\quad+
    \nabla_{\state}f_{\mathrm{other}}^\top
    \left(
        f(\state)+g(\state)\action
    \right).
\end{align}
Here, $f_p(\state)$ and $g_p(\state)$ denote the components of $f(\state)$ and $g(\state)$ corresponding to the position dynamics.

In addition to the CBF constraint, the MCBF controller employs a geodesic tangential guidance vector $\phi$ to encourage obstacle-circumventing motion. To avoid transient changes in guidance caused by dynamic obstacles, $\phi$ is computed using only environment-infrastructure obstacles. Specifically, $h^\phi$ is selected as the barrier function associated with the closest environment-infrastructure cluster obstructing the direct path between the robot and the target. Following \cite{xue_2025_mcbf}, let $e^{(0)}$ be one of $m$ uniformly sampled candidate directions satisfying $e^{(0)}\in\mathcal{N}(\nabla_{\state}h^\phi(\state))$. Each candidate generates a path $[\state^{(0)},\state^{(1)},\ldots,\state^{(N)}]$ along the isoline $h^\phi(\state)=h^\phi(\state^{(0)})$ using a first-order boundary approximation. The candidate with the minimum accumulated potential $C^{(N)}$ determines $\phi(\state)$ through its initial direction $e^{(0)}$. The state, direction, and accumulated potential are updated as
\begin{equation}
\label{eq:geo_approx}
\begin{aligned}
\state^{(i+1)}
&=
\state^{(i)}
+
\beta H(\state^{(i)})H(\state^{(i)})^\top \bm{e}^{(i)},\\
\bm{e}^{(i+1)}
&=
\frac{
H(\state^{(i)})H(\state^{(i)})^\top \bm{e}^{(i)}
}{
\left\|
H(\state)H(\state)^\top \bm{e}^{(i)}
\right\|_2
},\\
C^{(i+1)}
&=
C^{(i)}
+
\beta c(\state^{(i+1)},\bm{p}^{(xy)}_g).
\end{aligned}
\end{equation}
Here, $\beta$ is the step size approximated from the obstacle-cluster boundary length, $H(\state)$ contains an orthonormal basis of the tangent hyperplane to the corresponding isoline, and $c(\state,\bm{p}^{(xy)}_g)$ is a user-defined penalty function that evaluates the desirability of a state relative to the target position.

\subsubsection{Stage Selection}

The MCBF controller requires stage-dependent adaptation of both the CBF constraint and geodesic guidance. Conservative CBF settings improve dynamic-obstacle clearance but can impede narrow-passage traversal, while boundary-based geodesic guidance is necessary to capture narrow passages but can poorly represent the robot state in open space. We therefore define three collision-avoidance stages---dynamic-obstacle, narrow-space, and normal---and adapt the CBF parameters and geodesic guidance accordingly.

The dynamic-obstacle stage is selected if any dynamic obstacle $k$ satisfies
\begin{equation}
\nonumber
f^{(k)}_{\mathrm{dist}}(\bm{p}^{(xy)}_{\mathrm{rob}})
< r_{\mathrm{close}},\quad
\nabla_{\bm{p}_{c,k}}f^{(k)}_{\mathrm{dist}}(\bm{p}^{(xy)}_{\mathrm{rob}})^\top
\dot{\bm{p}}_{c,k}
< -v_{\mathrm{close}},
\end{equation}
where $v_{\mathrm{close}}>0$ is a predefined closing-speed threshold. These conditions identify a dynamic obstacle that is both within the specified proximity and approaching the robot at a rate requiring immediate avoidance.

If the dynamic-obstacle criterion is not satisfied, the narrow-space stage is evaluated using two complementary indicators derived from $\mathbb{P}_{\mathrm{env}}$. The inter-cluster indicator identifies narrow passages based on the minimum separation between distinct environment-infrastructure clusters. This criterion is inapplicable when the relevant boundaries are represented by a single cluster, as in enclosed regions with a narrow opening through which the robot must enter or exit. These configurations are termed \emph{confinement} and are identified using a separate confinement indicator. Dynamic obstacles in $\mathbb{P}_{\mathrm{dyn}}$ are excluded from both indicators, as their transient geometry is not used to characterize the underlying environment structure.

For the inter-cluster indicator, a rectangular region $\mathcal{R}$ of length $L$ and width $W$ is defined to capture the environment near the robot and along its current navigation direction. The region is centered at $\bm{p}^{(xy)}_{\mathrm{rob}}$ and oriented along
\begin{equation}
\boldsymbol{\varsigma}=
\begin{cases}
\phi(\state), & \text{if } \phi(\state) \text{ is defined},\\
[\cos\theta_{\mathrm{rob}},\sin\theta_{\mathrm{rob}}]^\top, & \text{otherwise},
\end{cases}
\end{equation}
where $\theta_{\mathrm{rob}}$ is the robot orientation. The minimum separation between distinct environment-infrastructure clusters within $\mathcal{R}$ is defined as
\begin{equation}
\nonumber
\resizebox{\columnwidth}{!}{$
r_{\mathrm{pass}}
=
\min\limits_{\substack{
\mathcal{P}_i^{(xy)},\,\mathcal{P}_j^{(xy)}\in\mathbb{P}_{\mathrm{env}}\\
i\neq j}}
\left\{
\left\|\bm{p}_i-\bm{p}_j\right\|_2
\;\middle|\;
\bm{p}_i\in\mathcal{P}_i^{(xy)}\cap\mathcal{R},\;
\bm{p}_j\in\mathcal{P}_j^{(xy)}\cap\mathcal{R}
\right\}.
$}
\end{equation}
The narrow-space stage is activated when $r_{\mathrm{pass}}<r_{\mathrm{narrow}}$.

The confinement indicator evaluates whether a point $\bm{p}^{(xy)}_{\mathrm{confine}}$ is enclosed by a single environment-infrastructure cluster with a narrow opening. Each point $\bm{p}^{(xy)}\in\mathbb{P}_{\mathrm{env}}$ is assigned to one of $N_{\mathrm{bin}}$ uniformly spaced bearing bins according to
\begin{equation}
\nonumber
\operatorname{bin}\!\left(\bm{p}^{(xy)}\right)
=
\left\lfloor
\frac{
\operatorname{atan2}\!\left(
p^{(y)}-p^{(y)}_{\mathrm{confine}},
p^{(x)}-p^{(x)}_{\mathrm{confine}}
\right)+\pi
}{
2\pi/N_{\mathrm{bin}}
}
\right\rfloor .
\end{equation}
Within each bearing bin, the environment-infrastructure point nearest to $\bm{p}^{(xy)}_{\mathrm{confine}}$ is retained, and its distance is denoted by $d_{\mathrm{bin}}$. A bin is blocked if a nearest point exists and $d_{\mathrm{bin}}<d_{\mathrm{confined}}$. The point $\bm{p}^{(xy)}_{\mathrm{confine}}$ is classified as confined when at least $90\%$ of the $N_{\mathrm{bin}}$ bins are blocked. When the robot is confined, adjacent bearing bins are examined for a discontinuity in $d_{\mathrm{bin}}$; a difference exceeding a predefined threshold identifies an opening in the cluster enclosing the robot.

The confinement test is first evaluated at the robot position, $\bm{p}^{(xy)}_{\mathrm{confine}}=\bm{p}^{(xy)}_{\mathrm{rob}}$. If the robot is confined, the narrow-space stage is activated and a temporary target is placed at the detected opening farthest from the robot to facilitate departure from the enclosure. Once the robot is no longer classified as confined, the temporary target is discarded and the original target $\bm{p}^{(xy)}_g$ is restored. The confinement test is then evaluated at the target position, $\bm{p}^{(xy)}_{\mathrm{confine}}=\bm{p}^{(xy)}_g$. If the robot is not confined but the target is, the narrow-space stage is activated when the robot is sufficiently close to $\bm{p}^{(xy)}_g$ to facilitate entry into the enclosure. Otherwise, the normal stage is selected.

\subsubsection{Stage-Dependent MCBF Adaptation}

The selected stage adapts the MCBF controller in two aspects: CBF constraint conservativeness and geodesic tangential guidance. First, the CBF conservativeness is adjusted through the class-$\mathcal{K}$ function $\alpha(h)$ and the safety margin $\delta$. The dynamic-obstacle, normal, and narrow-space stages use progressively less conservative settings:
\begin{equation}
\nonumber
\alpha_{\mathrm{dyn}}(h)
\leq
\alpha_{\mathrm{normal}}(h)
\leq
\alpha_{\mathrm{narrow}}(h),
\qquad
\delta_{\mathrm{narrow}}
\leq
\delta_{\mathrm{normal}}
\leq
\delta_{\mathrm{dyn}}.
\end{equation}

Second, the selected stage determines the isoline on which the geodesic tangential guidance vector $\phi$ is evaluated. In the dynamic-obstacle and normal stages, the geodesic approximation is initialized at the robot state,
\begin{equation}
\state^{(0)}=\state_{\mathrm{rob}},
\end{equation}
and therefore follows the isoline corresponding to the robot's current position. In the narrow-space stage, this isoline may exclude passages narrower than approximately $2h^\phi(\state_{\mathrm{rob}})$. The robot state is therefore projected onto the zero isoline, and the resulting projection $\state_{\mathrm{rob}}^{\mathrm{proj}}$ is used as the initial state:
\begin{equation}
\state^{(0)}=\state_{\mathrm{rob}}^{\mathrm{proj}},
\qquad
h^\phi(\state^{(0)})=0.
\end{equation}

These stage-dependent adaptations reduce CBF conservativeness in confined regions while preserving obstacle-circumventing geodesic guidance and increased clearance from rapidly approaching dynamic obstacles.

\begin{table*}[!t]
\centering
\setlength{\abovecaptionskip}{2pt}
\setlength{\belowcaptionskip}{2pt}
\caption{Comparison of the memory-aware perception-based MCBF, its ablated CBF variant, and baselines in target-reaching success rate, collision rate per human encountered, and trajectory tortuosity. $^{\dag}$MPC-CBF was infeasible at the initial position; tortuosity is undefined.}
\label{tab:simulation-results}

\setlength{\tabcolsep}{2.2pt}
\setlength{\aboverulesep}{0.2ex}
\setlength{\belowrulesep}{0.3ex}
\renewcommand{\arraystretch}{0.9}
\scriptsize

\resizebox{\textwidth}{!}{%
\begin{tabular}{@{}cc@{\hskip 6pt}ccc@{\hskip 6pt}ccc@{\hskip 6pt}ccc@{\hskip 6pt}ccc@{\hskip 6pt}ccc@{}}
\toprule
\multirow{2}{*}{\textbf{Env.}} & \multirow{2}{*}{\textbf{Sc.}}
& \multicolumn{3}{c}{\textbf{Adaptive MCBF (Ours)}}
& \multicolumn{3}{c}{\textbf{Adaptive CBF}}
& \multicolumn{3}{c}{\textbf{CE-CBF}}
& \multicolumn{3}{c}{\textbf{MPC-CBF}}
& \multicolumn{3}{c}{\textbf{DWA}} \\
\cmidrule(lr){3-5} \cmidrule(lr){6-8} \cmidrule(lr){9-11}
\cmidrule(lr){12-14} \cmidrule(lr){15-17}
& & Succ.\,\% & Coll./Enc. (\%) & Tort.
  & Succ.\,\% & Coll./Enc. (\%) & Tort.
  & Succ.\,\% & Coll./Enc. (\%) & Tort.
  & Succ.\,\% & Coll./Enc. (\%) & Tort.
  & Succ.\,\% & Coll./Enc. (\%) & Tort. \\
\midrule

& 1 & \textbf{100.0} & 4/56 (7.1) & 1.63
    & 0.0 & 5/60 (8.3) & 4.63
    & 0.0 & 9/55 (14.8) & 7.42
    & 0.0 & 5/29 (17.2) & 1.10
    & \textbf{100.0} & 4/56 (7.1) & 1.53 \\
& 2 & \textbf{100.0} & 3/32 (9.4) & 3.07
    & 0.0 & 0/0 & 129.53
    & 0.0 & 0/0 & 51.48
    & 0.0 & 0/0 & 1.10
    & 0.0 & 0/0 & 1.08 \\
& 3 & \textbf{100.0} & 1/20 (5.0) & 1.37
    & 40.0 & 1/39 (2.6) & 2.94
    & 0.0 &  4/60 (6.7) & 23.90
    & 0.0 & 3/24 (12.5) & 1.13
    & 60.0 & 1/17 (5.9) & 1.32 \\
& 4 & \textbf{100.0} & 2/27 (7.4) & 1.95
    & 40.0 & 2/33 (6.1) & 5.33
    & 0.0 & 0/0 & 99.26
    & 0.0 & 0/0 & 1.12
    & 0.0 & 0/3 (0.0) & 2.59 \\
& 5 & \textbf{100.0} & 0/14 (0.0) & 3.61
    & 80.0 & 6/50 (12.0) & 8.97
    & 0.0 & 2/15 (13.3) & 52.40
    & 0.0 & 0/0 & --$^{\dag}$
    & 0.0 & 2/8 (25.0) & 1.70 \\
\cellcolor{white}\multirow{-6}{*}{\textbf{Hospital}} & 6 & \textbf{100.0} & 8/45 (17.8) & 1.70
    & 20.0 & 9/79 (11.4) & 4.71
    & 0.0 & 1/67 (1.5) & 16.38
    & 0.0 & 1/1 (100.0) & 1.05
    & 80.0 & 2/50 (4.0) & 3.34 \\
% \rowcolor{TableOverall}
% \cellcolor{white}\multirow{-7}{*}{\textbf{Hospital}}
%   & \textbf{All} & \textbf{100.0} & 18/194 (9.3) & 2.22
%     & 30.0 & 23/261 (8.8) & 26.02
%     & 0.0 & 16/197 (8.1) & 41.81
%     & 0.0 & 9/54 (16.7) & 1.01
%     & 40.0 & 9/134 (6.7) & 1.93 \\

\midrule

& 1 & \textbf{100.0} & 0/11 (0.0) & 1.75
    & 80.0 & 0/10 (0.0) & 1.39
    & \textbf{100.0} & 1/13 (7.7) & 1.73
    & 0.0 & 1/9 (11.1) & 1.19
    & \textbf{100.0} & 0/6 (0.0) & 1.07 \\
\cellcolor{white}\multirow{-2}{*}{\textbf{Warehouse}}& 2 & \textbf{100.0} & 0/10 (0.0) & 2.28
    & 40.0 & 1/12 (8.3) & 4.89
    & 0.0 & 3/44 (6.8) & 23.60
    & 0.0 & 0/0 & --$^{\dag}$
    & \textbf{100.0} & 0/2 (0.0) & 1.19 \\
    
% \rowcolor{TableOverall}
% \cellcolor{white}\multirow{-3}{*}{\textbf{Warehouse}}
%   & \textbf{All} & \textbf{100.0} & 0/21 (0.0) & 2.02
%     & 60.0 & 1/22 (4.5) & 3.14
%     & 50.0 & 4/57 (7.0) & 12.67
%     & 0.0 & 1/9 (11.1) & 1.19
%     & \textbf{100.0} & 0/8 (0.0) & 1.13 \\
\midrule
\rowcolor{TableOverall}
\textbf{Summary}
  &  & \textbf{100.0} & 18/215 (8.4) & 2.17
    & 37.5 & 24/283 (8.5) & 20.30
    & 12.5 & 20/254 (7.9) & 34.52
    & 0.0 & 10/63 (15.9) & 1.12
    & 55.0 & 9/142 (6.4) & 1.73 \\
\bottomrule
\end{tabular}%
}
\vspace{-2mm}
\end{table*}

\section{Experiments}
\label{sec:experiments}
The proposed memory-aware multi-sensor perception system with MCBF-QP is evaluated in simulation and on hardware. The simulation study, presented in Section~\ref{sec:sim_experiments}, compares the proposed method with an ablated variant and three baseline methods to quantify its safety and liveness performance, while the hardware experiment, presented in Section~\ref{sec:hardware}, demonstrates deployment feasibility on a differential-drive Scout robot.

\subsection{Simulation Experiments}
\label{sec:sim_experiments}

\begin{figure}[!t]
    \centering
    \begin{subfigure}[t]{0.57\columnwidth}
        \centering
        \includegraphics[width=\linewidth]{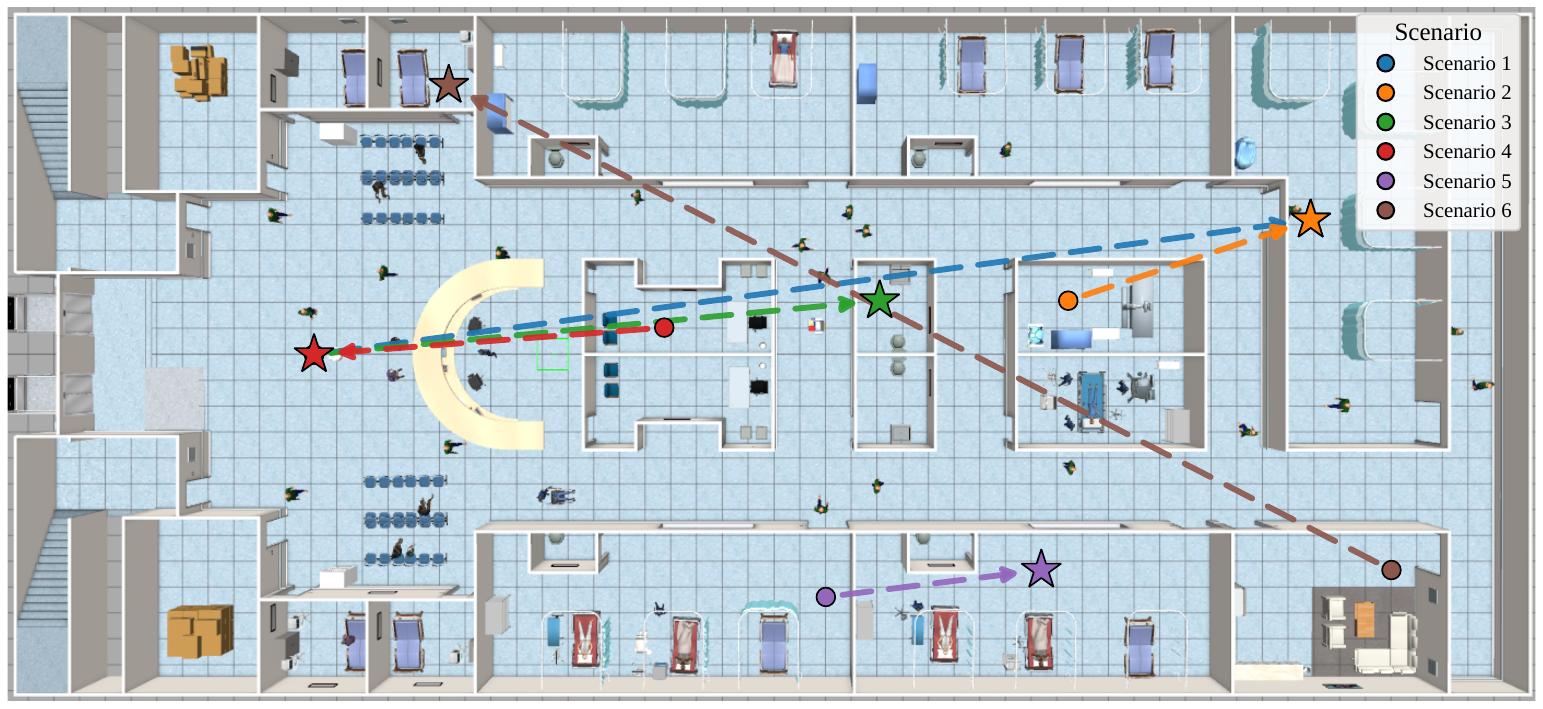}
        % \caption{Hospital}
        \label{fig:allscenarios_hospital}
    \end{subfigure}%
    \hfill
    \begin{subfigure}[t]{0.4\columnwidth}
        \centering
        \includegraphics[width=\linewidth]{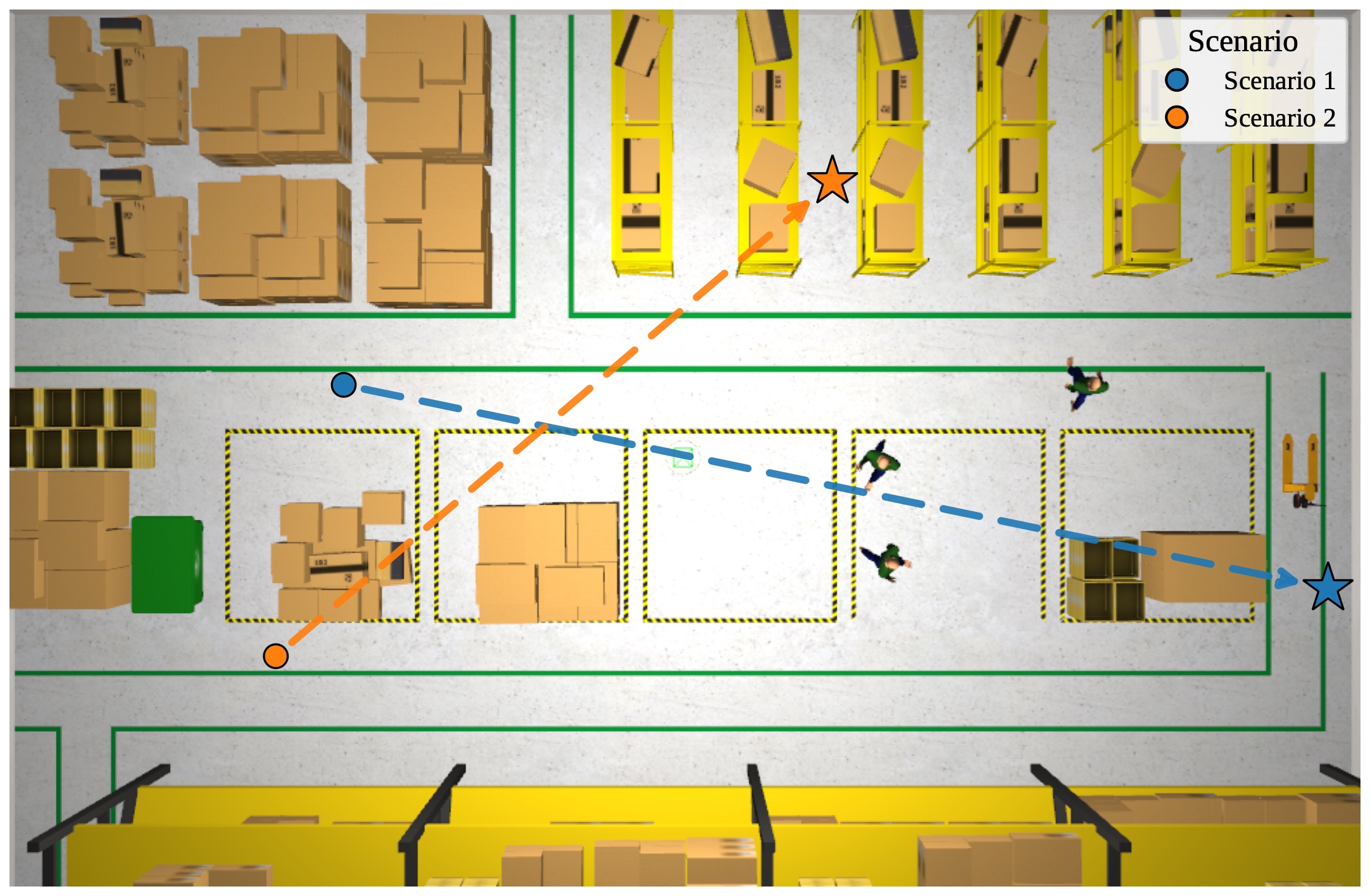}
        % \caption{Warehouse}
        \label{fig:allscenarios_warehouse}
    \end{subfigure}
    \vspace{-10pt}
    \caption{Hospital (left) and warehouse (right) simulation.}
    \vspace{-15pt}
    \label{fig:all-experiment-scenarios}
\end{figure}

\subsubsection{Experiment Design}
Two simulated environments are adapted from the AWS RoboMaker Warehouse World and Hospital World~\cite{aws_robomaker_worlds}. The robot is a Freight differential-drive base equipped with a $360^\circ$ 3D LiDAR and four RGB cameras, each with a $90^\circ$ field of view, providing full $360^\circ$ visual coverage. The camera-based dynamic obstacle detector is trained specifically for human detection; other movable obstacle classes can be incorporated by extending the training dataset with the corresponding categories. Six navigation scenarios are defined in the hospital environment and two in the smaller, less cluttered warehouse environment, with 20 and 3 human actors as dynamic obstacles per trial, respectively. Each scenario specifies fixed start and goal poses, with all scenarios shown in Fig.~\ref{fig:all-experiment-scenarios}. Each human follows an independently designed waypoint sequence without reacting to the robot and is initialized at a random position along its trajectory at the beginning of each trial, resulting in varying robot--human encounter timings. Each scenario is repeated five times for each baseline method to obtain performance measures that are less sensitive to the particular random initialization.

\subsubsection{Ablation and Baselines}
The proposed adaptive MCBF-QP is compared with an ablated variant, denoted \emph{Adaptive CBF}, which retains the same perception, obstacle representation, and parameter adaptation pipeline but replaces the MCBF-QP formulation with a standard CBF-QP.
Three existing perception-based navigation approaches are also considered: \emph{MPC-CBF}~\cite{mpc_cbf_ellipse_lidar}, which cluster LiDAR point clouds into fitted ellipses; \emph{DWA}~\cite{fox1997dynamic}, operating on a standard occupancy map; and \emph{CE-CBF}~\cite{cbf_circulation}, a CBF that learns local obstacle representations from LiDAR data using a GP and incorporates a circulation constraint to enhance liveness. \emph{MPC-CBF} and \emph{CE-CBF} do not retain previously observed environmental information, whereas \emph{DWA} uses a memory-aware occupancy map constructed through ray tracing.

\subsubsection{Metrics}
All methods are evaluated under identical scenarios and sensing conditions using success rate, human collision rate, and trajectory tortuosity. Success rate is measured as the percentage of trials in which the robot reaches the goal, while human collision rate is the proportion of human encounters resulting in a collision, with encounters determined by the ground-truth robot--human distance being within three meters. Trajectory tortuosity is defined as the traveled path length divided by the straight-line distance between the initial and final positions, with larger values indicating greater oscillation and detouring within a scenario.

\begin{figure}[!t]
    \centering
    \includegraphics[width=\columnwidth, trim={0 0mm 0 4mm}]{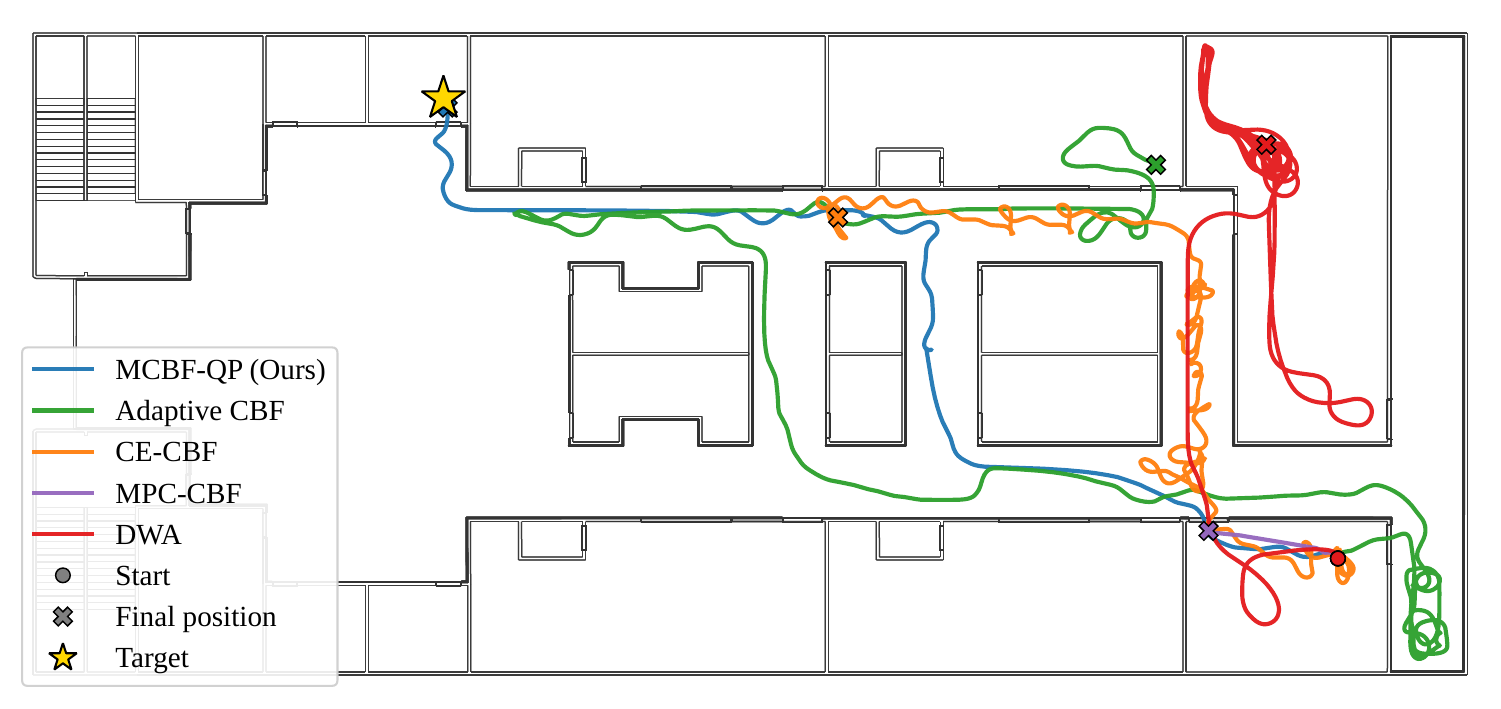}
    \caption{Representative navigation trajectories of the evaluated controllers in Scenario~6 of the hospital environment.}
    \label{fig:scenario6-path-overlays}
    \vspace{-15pt}
\end{figure}

\subsubsection{Result Analysis}
Table~\ref{tab:simulation-results} shows that the proposed MCBF-QP is the only method to achieve a 100\% success rate while maintaining a comparable human collision rate. The memory-unaware \emph{CE-CBF} and \emph{MPC-CBF}, which construct obstacle representations solely from current observations without retaining previously observed environmental information, achieve the lowest success rates ($<12.5\%$). \emph{MPC-CBF} approximates LiDAR observations with fitted ellipses for real-time control, but the conservative approximation frequently renders the controller infeasible. \emph{CE-CBF} introduces a circulation constraint for liveness, but its locally computed guidance is sensitive to LiDAR point-cloud distributions and can provide inconsistent directions that hinder target convergence. These results indicate that liveness enhancement alone is insufficient without persistent environmental information.

The memory-aware \emph{Adaptive CBF} and \emph{DWA} achieve higher success rates ($30\%-50\%$) by retaining previously observed environmental information, but neither incorporates a liveness-enhancement strategy. Consequently, they remain susceptible to local undesirable equilibria and limit cycles, indicating that persistent environmental information alone is also insufficient for reliable target convergence.

Across the baselines, dynamic obstacles can further push the robot away from the target, resulting in long detours and high trajectory tortuosity even when the target is reached, as illustrated in Fig.~\ref{fig:scenario6-path-overlays}. The proposed method combines persistent infrastructure mapping of explored regions with a modified MCBF-QP that uses the perceived environment to adapt the guiding direction and parameters, enabling efficient convergence toward the target.

\subsection{Hardware Experiment}
\label{sec:hardware}

%--------------------------------------------------------------------------
\subsubsection{Platform and Computing Architecture}
% \label{sec:hardware}

The hardware platform consists of a Scout~2.0 skid-steer mobile base equipped with a rear-mounted SICK multiScan136 3D LiDAR, a front-mounted Intel RealSense D435 RGB-D camera to provide additional depth observations in the near-field where LiDAR coverage is sparse, and a rear-mounted Logitech C920 RGB camera for human detection. All sensors are rigidly mounted with fixed extrinsic calibration. The complete system pipeline runs onboard an NVIDIA Jetson AGX Orin.

%--------------------------------------------------------------------------
\subsubsection{Navigation Experiment}
\label{sec:hw_navigation}

% The system was evaluated in an unmapped indoor lobby containing narrow passages, concave structures, and dynamic obstacles. The robot constructed the occupancy map and distance fields online and successfully reached the goal. Fig.~\ref{fig:hw-scenarios} overlays the onboard-estimated trajectory on the resulting occupancy map, demonstrating the end-to-end feasibility of the proposed system using only onboard sensing and computation.

\begin{figure}[!t]
    \centering
    \begin{subfigure}[b]{0.52\linewidth}
        \centering
        \includegraphics[width=\linewidth]{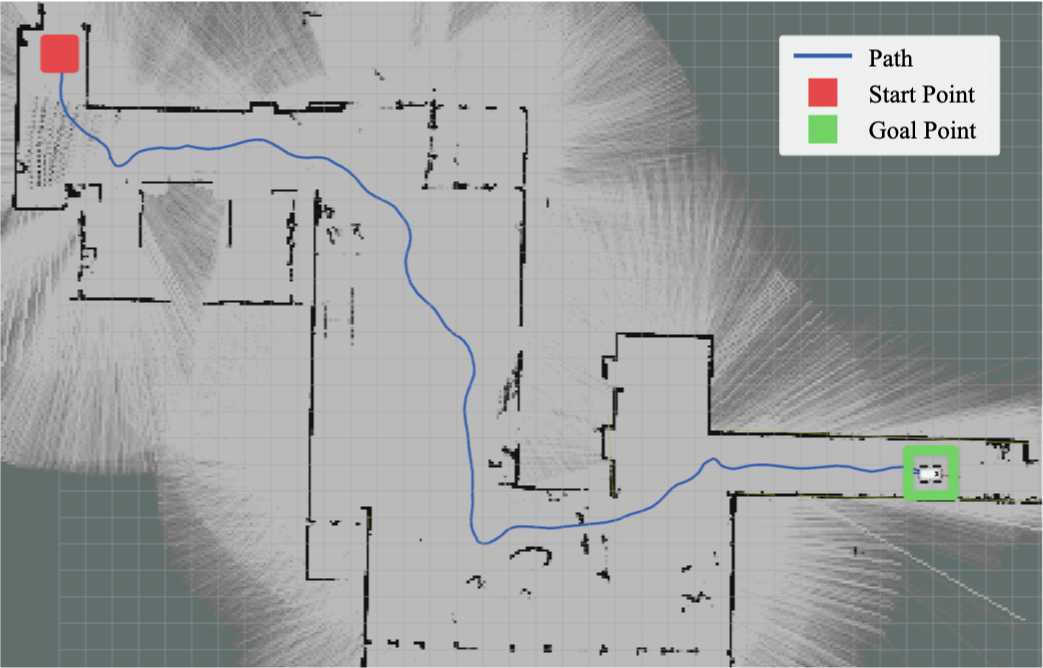}
        % \caption{Indoor.}
        \label{fig:hw-indoor}
    \end{subfigure}
    \hfill
    \begin{subfigure}[b]{0.458\linewidth}
        \centering
        \includegraphics[width=\linewidth]{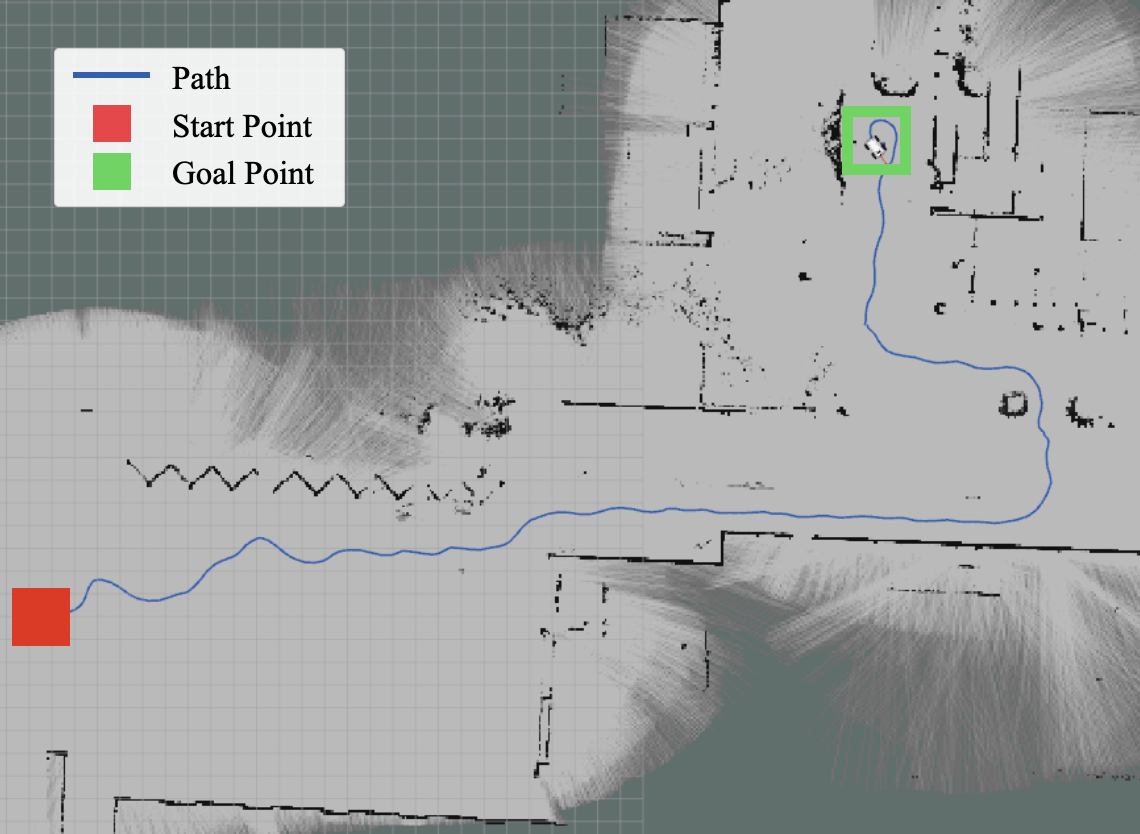}
        % \caption{Outdoor.}
        \label{fig:hw-outdoor}
    \end{subfigure}
    \vspace{-15pt}
    \caption{Robot trajectories and online occupancy maps for indoor (left) and outdoor (right) hardware experiments.}
    \vspace{-15pt}
    \label{fig:hw-scenarios}
\end{figure}

We qualitatively evaluated the complete system in two previously unmapped real-world environments. The indoor lobby and outdoor courtyard both contain narrow passages, concave structures, moving pedestrians, sloped terrain, and irregular structures or vegetation. In both environments, the robot constructed the occupancy map and distance fields online and successfully reached the goal. Fig.~\ref{fig:hw-scenarios} overlays the onboard-estimated trajectory on the resulting occupancy map, demonstrating the end-to-end feasibility of the proposed system using only onboard sensing and computation.

\section{Discussions and Future Work}
This work presented a memory-aware, multi-sensor navigation framework that combines an online persistent environmental representation with stage-adaptive MCBF-QP control to enable safety and liveness for navigation in previously unmapped, dynamic indoor/outdoor environments. Future work could investigate the increasing computational burden of processing larger occupancy maps, which can slow controller execution under limited onboard computing resources, potentially through adaptive memory management and persistent representation. Another direction is to further improve dynamic obstacle avoidance by incorporating accurate obstacle motion prediction, with the goal of achieving collision-free navigation in complex and crowded environments.

\bibliographystyle{IEEEtran}
\bibliography{main}
\end{document}